# Cancer image classification based on DenseNet model

**Ziliang Zhong[1], Muhang Zheng[1], Huafeng Mai[2], Jianan Zhao[3] and Xinyi Liu[4]**

[1] New York University Shanghai, Shanghai, zz1706@nyu.edu, China

[1] South China Agricultural University, Shenzhen, 1315866130@qq.com, China

[2] University of Arizona, Tucson, huafengmai@email.arizona.edu, United States

[3] University of California, La Jolla, jiz038@ucsd.edu, United States

[4] University of California, San Diego, La Jolla, xil005@ucsd.edu, United States

[*]Corresponding author's e-mail: 1315866130@qq.com

**Abstract.** Computer-aided diagnosis establishes methods for robust assessment of medical image-based examination. Image processing introduced a promising strategy to facilitate disease classification and detection while diminishing unnecessary expenses. In this paper, we propose a novel metastatic cancer image classification model based on DenseNet Block, which can effectively identify metastatic cancer in small image patches taken from larger digital pathology scans. We evaluate the proposed approach to the slightly modified version of the PatchCamelyon (PCam) benchmark dataset. The dataset is the slightly modified version of the PatchCamelyon (PCam) benchmark dataset provided by Kaggle competition, which packs the clinically-relevant task of metastasis detection into a straight-forward binary image classification task. The experiments indicated that our model outperformed other classical methods like Resnet34, Vgg19. Moreover, we also conducted data augmentation experiment and study the relationship between Batches processed and loss value during the training and validation process.

**1. Introduction**
With the rapid development of science and technology, computer-aided diagnosis (CAD) establishes methods for robust assessment of medical image-based examination. In recent years, image processing introduced a promising strategy to facilitate disease classification and detection while diminishing unnecessary expenses. Traditional image processing and machine learning techniques require extensive pre-processing, manual extraction of visual features, and handcrafted segmentation before classification. However, such kind of feature-based work is labor-intensive. Deep neural network models are of growing interest for their capacity to automatically generate useful low dimensional representations by utilization of hierarchical feature extraction and classification, which achieve high accuracy on image process and cancer disease diagnosis.

Image classification is a fundamental and essential computer vision task. In recent years, convolutional neural networks (CNNs) have become the dominant machine learning approach for computer vision recognition. The original LeNet5 [14] consisted of 5 layers, VGG featured 19 [5], and Residual Networks (ResNet) [4] have surpassed the 100-layer barrier. However, there are potential problems with these models, such as too many parameters, gradient disappearance and difficulty in training. Dense Convolutional Network (DenseNet) [3] has dense connectivity, which is superior to other models like Vgg and Resnet. A different connectivity pattern of DenseNet model from other CNN is direct connections from any layer to all subsequent layers, which can further improve the information flow between layers. Therefore, DenseNet can alleviate the gradient vanishing problem, enhance the propagation of feature maps, and reduce some parameters very well.

So, in this paper, we deal with the problem of these metastatic cancer image diagnosis as an image classification task in computer vision processing. For example, as we can see from the example given in Figure 1, the images are small image patches taken from larger digital pathology scans, and each input image has a binary label 0 or 1 (Lable 0 means not cancer, label 1 means cancer). We utilize DenseNet Block to identify metastatic cancer in small image patches taken from larger digital pathology scans. We are the first to apply DenseNet model to solve such a medical image problem. We conduct experiments to evaluate the DenseNet based approach on the slightly modified version of the PatchCamelyon (PCam) benchmark dataset provided by the Kaggle competition. The dataset is the slightly modified version of the Pcam benchmark dataset provided by Kaggle competition, which packs the clinically-relevant task of metastasis detection into a straight-forward binary image classification task. During the experiment, we use data augmentation techniques to get better results. The experiments show that our model outperformed the other classical methods like Resnet34, Vgg19. Specifically, our DenseNet201 (TTA) model is 2.37% higher in Auc-Roc score and 2.4% higher in Accuracy metric than Vgg19 model. Experiments show our idea using DenseNet model is effective for cancer image classification tasks.

The main contributions of this paper are presented as follows:
- To the best of our knowledge, this is a new application innovation to identify cancer by DenseNet image method;
- We evaluate our method DenseNet201 on modified Pcam datasets. And experiments show that our model achieves superior performance over the comparison approaches;
- We conducted data augmentation experiment on DenseNet201(TTA) model. The experiment indicates that our model has the highest Auc-Roc score and Accuracy score.

The rest of this paper is organized as follows. Related work and recent research progress on cancer image detection and image classification are introduced in Section II. Section III describes our algorithm based on DenseNet Block, a convolutional neural network. Next, in Section IV, we conduct experiments between the proposed model and other classic convolutional neural network models for image classification. Finally, the conclusion of this paper is in Section V.

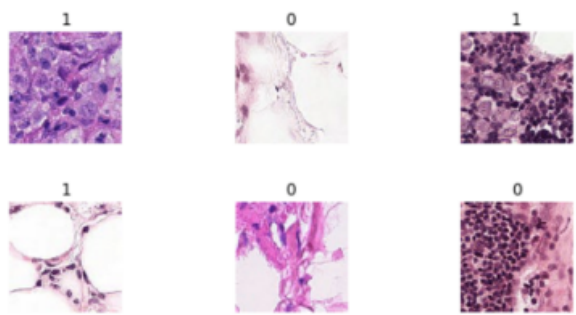

**Figure 1.** Examples of input image.

An example to show metastatic cancer in small image patches taken from larger digital pathology scans. Lable 0 means the image sample is cancer, and label 1 means the image sample is not cancer

## 2. Related work

In medical science, patients usually are initially used clinical screening, followed by histopathological analysis of cancer sites to make a preliminary diagnosis. Automatic classification of cancer using histopathological images is a difficult task for accurately detecting cancer, especially for identifying metastatic cancer in small image patches taken from larger digital pathology scans. Computer-aided diagnosis can simplify this process, and may be more reliable and economical. Image processing introduced a promising strategy to facilitate disease classification and detection, while diminishing unnecessary expenses.

A large body of works has been published about cancer detection using various image processing and machine learning techniques ([12], [13]). Application of these traditional approaches is limited due to manual feature extraction of the specific features and labor intensive. Deep learning methods offer automated, sensitive and accurate models to feature extraction from medical image data.

Due to the establishment of large-scale hand-labeled data sets (such as ImageNet [6]), the performance of image classification has recently developed rapidly. Many efforts have been dedicated to extending deep convolutional networks for medical image classification. Wang et al. [7] applied VGG16 deep network to identify breast cancer. Esteva et. al. [8] conducted a study on skin cancer detection using Inception V3, which was done to classify malignancy status. And Habibzadeh et al. [9] use ResNet model for Breast Cancer Histopathological Image Classification. In addition, many studies showed that, compared with conventional machine learning, deep learning techniques are continuously being applied to medical image data diagnosis and improve the performance such as [2], [10], and [11].

This research proposes the combination of deep learning thought with DenseNet architecture to to distinguish the small metastatic cancer image patches taken from larger digital pathology scans. We show a highly accurate automated framework that can be used for cancer detection classification. Our framework also uses the data augmentation technique for advanced pre-processing.

**3. Methodology**

Dense Convolutional Network (DenseNet) [3] has dense connectivity compared to other models such as Vgg [5] and Resnet [4]. DenseNet can alleviate the vanishing-gradient problem, enhance the propagation of feature maps, and reduces the number of parameters.

The overall illustration of the architecture of the proposed model is shown in Figure 2. In DenseNet model, a different connectivity pattern with other CNN is direct connections from any layer to all subsequent layers, which can further improve the information flow between layers. Hence, the l-th layer receives the feature-maps of all preceding layers, and the formula is computed as follows:

$$x^l = H^l\left([x^0, x^1, \ldots, x^{l-1}]\right) \qquad (1)$$

where l indexes the layer, $x^l$ denotes the output of the l-th layer. $[x^0, x^1, \ldots, x^{l-1}]$ means the concatenation of the feature-maps produced in layers $0, 1, 2, \ldots, l-1$. And $H^l$ can be a composite function of operations such as Rectified Linear Units (ReLU), Pooling, Convolution(Conv), or Batch Normalization (BN) [16]. The most significant difference between the DenseNet model and the ResNet model is as follows. ResNet adds a skip-connection that bypasses the non-linear transformations with an identity function:

$$x^l = H^l(x^l) + x^{l-1} \qquad (2)$$

The model DenseNet201 (TTA) denotes that we use DenseNet201 model to training set, but the test set is augmented. Our main idea is to expand the test dataset through data augmentation, and then multiple predictions to improve accuracy. Data Augmentation can increase the diversity of samples by making minor changes such as random resizing, rotating, cropping, and flipping the original image. Data augmentation is essential step to have enough diversed samples and learn a deep network from the images [1].

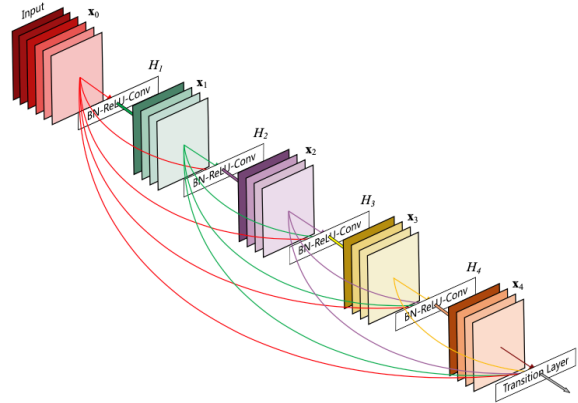

**Figure 2.** The architecture of DenseNet model.

## 4. Experiments

*4.1. Experimental Data and Experimental Settings*

In this section, we trained and validated our model on a modified version of the PatchCamelyon (PCam) dataset. PCam packs the clinically-relevant task of metastasis detection into a straight-forward binary image classification task. The data for this paper is a slightly modified version of the PatchCamelyon (PCam) benchmark dataset provided by Kaggle competition. The original PCam dataset contains duplicate images due to its probabilistic sampling; however, the modified version does not contain duplicates. Now, the modified PCam dataset contains 220025 image samples, including 89117 positive samples (with cancer) and 130908 negative samples (without cancer).

The dataset is randomly divided into two parts, with the proportion of 8:2. Four folds are used for training, and one fold is used for testing. For each image, the dimension of the input matrix is 96x96. During training, the batch size and learning ratio are set to 64 and 0.0001, respectively. We minimize the loss function using Adam optimizer [15]. Moreover, we implement our model on PyTorch with a GPU.

To evaluate the performance of the DenseNet model, we employ two types of indicators: Auc-Roc score and Accuracy score. The Auc-Roc score metric is the area under roc curve, which is known as the receiver operating characteristic curve. The Auc-Roc score can provide a comprehensive evaluation of the performance of the model, the higher scores indicating a better model. The roc curve is drawn by plotting the true positive rate (*TPR*) against the false positive rate (FPR) under various threshold settings. The calculation formulas for TPR and FPR are calculated as follows:

$$TPR = \frac{TP}{TP + FN} \quad (3)$$

$$FPR = \frac{FP}{FP + TN} \quad (4)$$

where TP and FN denote the number of positive samples that are predicted correctly and incorrectly, respectively. TN and FP mean the number of negative samples that are classified truly and falsely, respectively.

*4.2. Experimental Results and Analyses*

For our task, to evaluate the performance of our proposed model comprehensively, we select some previous classical studies, and compare them using the same dataset and evaluation metrics. The following several models are considered as the benchmarks for our experiments: Resnet34 and Vgg19.

Our model variants are DenseNet201 and DenseNet201 (TTA), which are compared with Resnet34 model and Vgg19 model on Pcam dataset. Table 1 shows the experimental results of competing models. It is easy to find out that our DenseNet based model outperforms the other two models on both Auc-Roc score and Accuracy metric. DenseNet201 (TTA) model has the highest value in both Auc-Roc Score and Accuracy metric, and DenseNet201 model is the second. DenseNet201 (TTA) model is 2.37% higher in Auc-Roc score and 2.4% higher in Accuracy metric than Vgg19 model. Therefore, the experimental results demonstrate that our model is effective for cancer image classification task.

**Table 1.** The results of different models on Pcam datasets in cancer image classification.

| Models | Auc Roc Score | Accuracy |
| --- | --- | --- |
| Resnet34 | 0.9633 | 0.975 |
| Vgg19 | 0.9473 | 0.965 |
| **DenseNet201** | 0.965 | 0.980 |
| **DenseNet201(TTA)** | **0.971** | **0.989** |

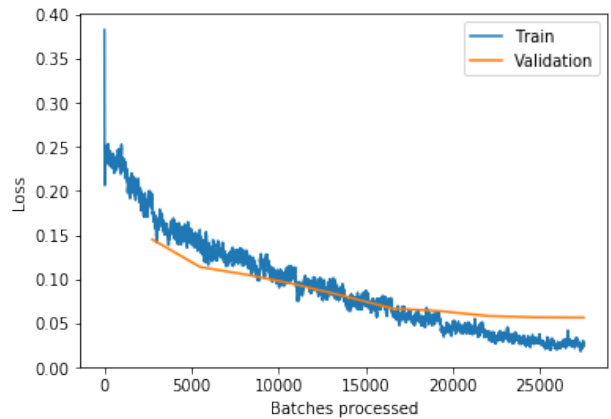

**Figure 3.** The relationship between Batches processed and loss value during the training and validation process.

Figure 3 shows the relationship between Batches processed and loss value during the training and validation process. In the beginning, with the increase of Batches processed, loss value has a significant downward trend in both training and validation process. However, when the Batches processed becomes very large, the decreasing trend of Loss value becomes gradually flattening. The reason maight be that when the number of iterative training of the model increases, the performance of the model becomes better, but when it reaches a certain number of times, the performance of the model tends to stabilize. At this time, more training does not necessarily guarantee better results but wastes time and computing resources. So, proper Batches processed is good for model training. These results can be constructive to guide the direction of future work about model training.

## 5. Conclusions

In this paper, we design a novel DenseNet based model for metastatic cancer image classification. Our model employs DenseNet Block to effectively capture the important characteristic information of metastatic cancer in small image patches taken from larger digital pathology scans. Experiments on Pcam dataset show that our model achieves superior performance over state-of-the-art approaches. In the comparison model, Our model has the best Auc-Roc score and Accuracy value relative to the compared models. In the future, we will improve the DenseNet based model for metastatic cancer image classification, obtain higher metric scores, and more accurate recognition.

**Acknowledgments**
The author Ziliang Zhong and Zheng Muhang thanks Huafeng Mai for the help in experiments, thanks Jianan Zhao for the contribution of the idea. In addition, we also thanks Xinyi Liu for fruitful discussions. Finally, we sincerely thanks to the free resources provided by Kaggle platform.